\documentclass{article}
\usepackage[utf8]{inputenc}
\usepackage{setspace}
\usepackage{geometry}
\usepackage{amsmath,amssymb,amsfonts}
\usepackage{algorithm}  
\usepackage{algpseudocode} 
\usepackage{graphicx}
\usepackage{url}
\usepackage[figuresright]{rotating}
\usepackage{array}
\usepackage{adjustbox}
\usepackage{longtable}
\title{{Dataset Generation for Drone Optimal Placement Using Machine Learning}
\date{2022\\ November}
\author{Jialin Hao  \thanks{This work is part of the Ph.D work funded by DigiCosme project: ANR11LABEX0045DIGICOSME, supervised by Djamal Zeghlache, Télécom SudParis and Rola Naja, ECE Paris}, Télécom SudParis, Institut Polytechnique de Paris 
}}
\begin{document}
\maketitle

\tableofcontents
\newpage

\listoffigures
\listoftables
\newpage

Unmanned aerial vehicle (UAV), or drone is increasingly becoming a promising tool in communication system. 
This report explains the generation details of a dataset which will be used to designing an algorithm for the optimal placement of UAVs in the drone-assisted vehicular network (DAVN) \cite{art:45}. The goal is to improve the drones' communication and energy efficiency after our previous work \cite{art:hjl-2022}. The report is organized as followed: the first section is devoted to the delay analysis of the vehicle requests in the DAVN using queuing theory; the second part of the report models the energy consumption of the drones while the third section explains the simulation scenario and dataset features. The notations and terminologies used in this report are summarized in the last section.

\section{End to End Delay Analysis}
\begin{figure}[htbp]
    \centering
    \includegraphics[width=\linewidth]{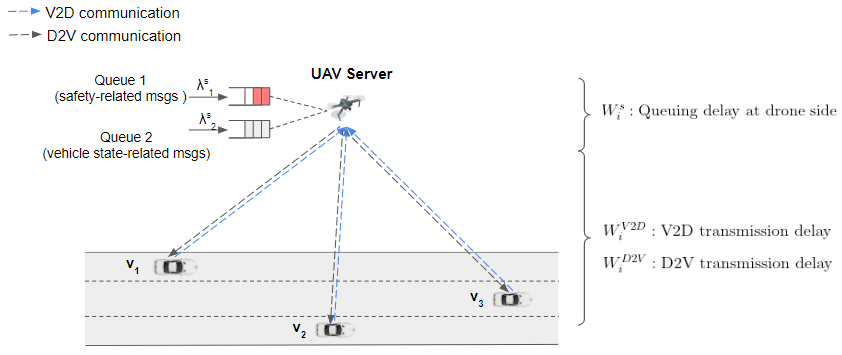}
    \caption{A simple DAVN with V2D and D2V communications.}
    \label{fig:DAVNQueue}
\end{figure}

This section analyzes the vehicle-to-drone (V2D) communication delay with the help of queuing theory. In fact, the total V2D delay of a class-$i$ request from a vehicle-$j$, denoted as $E[W_{i,j}]$, consists of three parts:
\begin{itemize}
    \item V2D propagation delay, $W_{i,j}^{V2D}$,
    \item queuing delay at the drone side, $W_{i,j}^s$,
    \item and drone-to-vehicle (D2V) propagation delay, $W_{i,j}^{D2V}$.
\end{itemize}
Thus, we have
\begin{equation}
   E[W_{i,j}] = 
   E[W_{i,j}^{V2D}] + E[W_{i,j}^{s}] + 
   E[W_{i,j}^{D2V}] \leq T_i, \forall j \in \mathbb{V},
   \label{eq:delay}
\end{equation}
where $\mathbb{V}$ is the set of vehicles in the scenario. The transmission delays at both the vehicle and the UAV side are neglected since these values are very small compared to the propagation delays. It should be noted that if a request of class-$i$ is not responded by the UAV after a certain time, $T_i$, where $T_i$ is the maximum waiting time for request class-$i$, the request will be regarded as expired and will not be processed by the UAV later\\
Fig.~\ref{fig:DAVNQueue} illustrates the queuing model for a single-drone-assisted vehicular network. The V2D and D2V communications are represented by the blue and grey dashed lines. At the mean time, the queuing model at UAV side is also highlighted. In the following sections, we will tackle the three delays in detail. 

\subsection{V2D and D2V Propagation Delay}
The V2D and D2V propagation delays of a class-$i$ request are calculated by
\begin{equation}
    \begin{aligned}
    & E[W_{i,j}^{V2D}] = \frac{d_{i,j}}{r_i}, \\ 
    & E[W_{i,j}^{D2V}] = \frac{d^{'}_{i,j}}{r_i}, 
    \end{aligned}
\end{equation}
where $d_{i,j}$ and $d^{'}_{i,j}$ are the distances between the UAV and the vehicle-$j$ from which the request is sent during the V2D and D2V communication, $r_i$ is the achievable propagation rate.

\subsection{Queuing Delay at the Drone's Side}
Inside of this section, we relief the server index $s$ and vehicle index $j$ in the expressions for simplicity, since the delay should be the same for any vehicle in the scenario. However, outside this section, the two indexes are kept.\\
We consider two types of message that can be sent by the vehicles to the drone, one is safety-related message with high priority, the other is vehicle state-related message with low priority. When the vehicle faces a road risk, or an emergency vehicle, it will send a safety message to the corresponding drone. On the other hand, each vehicle update its state to the drone by periodically sending a vehicle state information containing vehicle's GPS position and kinematic parameters. It should be noticed that the length of safety-related messages varies and is assumed to follow an exponential distribution. On the contrary, the length of vehicle state information is a constant and never changes.\\ 
At the drone's side, the service of a low priority request (i.e. vehicle state information) can be interrupted by the arrival of a high priority request (i.e. safety message). That is to say, the queuing model is a priority queuing with preemption.
In this case, the waiting times of the high priority requests, denoted as $W_1$, are not affected by the low priority requests, and are only related to the arriving process and service process of requests of class-$1$.
On the other hand, the waiting times of low priority requests, denoted as $W_2$, are affected by high priority requests, with an additional waiting time due to the arrival and interruption from a high priority request. 

\subsubsection{Definitions and notations}
We model the queuing delay of the safety message at the drones as a M/G/1 queue. The arriving process follows a Poisson process while the service time follows an exponential distribution. Moreover, the service times for different safety messages are independent and identically distributed. Note that for stability, it is required that the occupation rate $\rho _i = \lambda _i E[B _i]$ is less than one. The arrival process and service time distributions for safety-related messages and vehicle state-related messages are shown in Table. \ref{tab:Qprocess}\\
\begin{table}[htbp]\small
\centering
\caption{The arrival process and service time distributions}
\vspace{8pt}
\label{tab:Qprocess}
\renewcommand\arraystretch{1.3}
\begin{tabular}{|c|c|c|}
\hline
 & Arrival Process & Service Time Distribution \\ \hline
Safety Message & Poisson & Exponential \\ \hline
Vehicle State Information & Uniform & Constant \\ \hline
\end{tabular}%
\end{table}

We define the following variables for the drone:
\begin{itemize}
    \item $\rho _i$: the occupation rate of a class-$i$ request;
    \item $\lambda _i$: the arrival rate of a class-$i$ request;
    \item $E[B _i]$: the mean service time at the drone of a class-$i$ request;
    \item $E[W _i]$: the mean waiting time in the queue of a class-$i$ request;
    \item $E[R _i]$: the mean residual service time of a class-$i$ request;
    \item $E[S _i]$: the mean sojourn time of a class-$i$ request, note that $E[S _i] = E[B _i] + E[W _i]$;
    \item $E[L _i]$: the average number of requests of class $i$ waiting in the queue.
\end{itemize}

Thus, for the total incoming traffic at the drone side, we have the following:
\begin{equation}
    \lambda  = \sum ^n_{i=1} \lambda _i
    \label{eq:lambda}
\end{equation}
\begin{equation}
    E[B ] = \sum ^n_{i=1} \frac{\lambda _i}{\lambda } \cdot E[B _i]
    \label{eq:EB}
\end{equation}
\begin{equation}
    \rho  = \lambda  \cdot E[B ]
    \label{eq:rho}
\end{equation}

\subsubsection{Mean waiting time of high priority requests}
The average of $W _1$ can be expressed as followed:
    \begin{equation}
        E[W _1] =  E[L _1]E[B _1] + \rho _1 E[R _1],
    \end{equation}
where $L _1$ denote the number of high priority request waiting in the queue.\\
According to Little's law we have
\begin{equation}
    E[L _1] = \lambda _1 E[W _1].
\end{equation}
Combining the two equations yields
\begin{equation}
    E[W _1] = \frac{\rho _1 E[R _1]}{1- \rho _1}.
    \label{eq:EW1}
\end{equation}
Since we have
\begin{equation}
    E[R _1] = \frac{E[B _1^2]}{2E[B _1]},
\end{equation}
Equation \eqref{eq:EW1} becomes
\begin{equation}
     E[W _1] = \frac{\rho _1}{2(1- \rho _1)} \cdot \frac{E[B _1^2]}{E[B _1]}.
\end{equation}
The sojourn time is then
\begin{equation}
    E[S _1] = E[W _1] + E[B _1] = \frac{\rho _1}{2(1- \rho _1)} \cdot \frac{E[B _1^2]}{E[B _1]} + E[B _1].
\end{equation}

\subsubsection{Mean waiting time of low priority requests} 
As explained before, the waiting time of low priority requests can be expressed as 
$$E[W _2] = E[B _2] + E[W _+]$$
The customer has to first wait for the sum of the service times of all customers with the same or higher priority present in the queue plus the remaining service time of the customer in service. So
\begin{equation}
    E[B _2] = \sum_{j=1}^{2} E[L _j]E[B _j] + \sum_{j=1}^2 \rho _j E[R _j].
\end{equation}
On the other hand, the $W _+$ is related to all the higher priority requests arriving during its waiting time and service time. This leads to
\begin{equation}
    E[W _+] = \lambda _1 E[W _1] E[B _1].
\end{equation}
Applying Little's law
$$E[L _2] = \lambda _2 E[W _2],$$ 
we have
\begin{equation}
    E[W _2] = \frac{\sum _{j=1}^2 \rho _j E[R _j]}{\left( 1- \left( \rho _1 + \rho _2 \right) \right) \left( 1- \rho _1 \right)}.
\end{equation}
The mean sojourn time $E[S_2]$ of a class-$i$ customer follows from $E[S_2] = E[W_2] + E[B_2]$, yeilding
\begin{equation}
    E[S_2] = \frac{\sum _{j=1}^2 \rho_j E[R_j]}{\left( 1- \left( \rho_1 + \rho_2 \right) \right) \left( 1- \rho_1 \right)} + E[B_2].
    \label{ES2}
\end{equation}
Since we have
\begin{equation}
    E[R_i] = \frac{E[B_i^2]}{2E[B_i]},
\end{equation}
Equation \eqref{ES2} finally becomes
\begin{equation}
    E[S_2] = \frac{1}{\left( 1- \left( \rho_1 + \rho_2 \right) \right) \left( 1- \rho_1 \right)} \cdot \sum _{j=1}^2 \rho_j \frac{E[B_i^2]}{2E[B_i]} + E[B_2].
    \label{ES2}
\end{equation}

\section{Energy Consumption Model}
The energy consumed by the UAV-$i$ to perform a complete data transfer is composed of two components: the communication energy that is used for the data transfer from the UAV to vehicles and other UAVs, $E^c_i$, and the propulsion energy of the UAV to adjust its location for data transfer, $E^m_i$, as in the following equations:
\begin{equation}
    \begin{split}
     E_i &= E^{c}_i + E^{m}_i \\
       &= \sum_{j \in \mathbb{V}}E^{D2V}_{i,j} + \sum_{m \in \mathbb{U}}E^{D2D}_{i,m} + E^{m},
    \end{split}
    \label{eq:energyconsumption}
\end{equation}
where $\mathbb{V}$ is the set of all vehicles in the communication range of the UAV-$i$, $\mathbb{U}$ denotes the set of all UAVs.
One can see that we consider both the drone-to-vehicle (D2V) and the drone-to-drone (D2D) communications. In the following sections, each of the components will be analysed.

\subsection{Energy for communications}
\begin{figure}[htbp]
    \centering
    \includegraphics{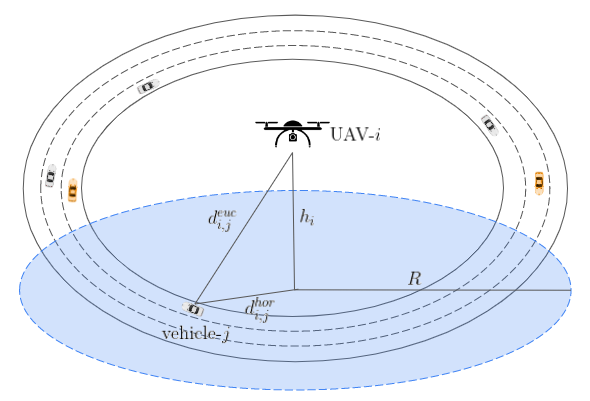}
    \caption{Scenario where the UAV-$i$ is communicating with vehicle-$j$}
    \label{fig:computationscenario}
\end{figure}

In order to model the communication energy consumption of our scenario, we first look at a simple scenario where the UAV-$i$ is communicating with vehicle-$j$, as illustrated in Fig. \ref{fig:computationscenario}. $h_i$ is the height of the UAV-$i$, $R$ is the communication range of the UAVs. $d_{i,j}^{euc}$ and $d_{i,j}^{hor}$ denote the Euclidean distance and the horizontal distance between the UAV-$i$ and vehicle-$j$, respectively. 

\paragraph{Drone-to-vehicle path loss}
The D2V path loss is also known as air-to-ground path loss. In \cite{art:atgloss}, the probability of having a line-of-sight link between the UAV-$i$ and vehicle-$j$ is formulated as followed:
\begin{equation}
    P^{D2V}_{i,j}(LoS) = \frac{1}{1+a \ \text{exp} \left( -b \left(\theta_{i,j} -a \right) \right)}
    \label{eq:plos}
\end{equation}
where $a$ and $b$ are environmental constant depending
on rural or urban areas. $\theta$ is the elevation angle between UAV-$i$ and vehicle-$j$, and it is equal to $\text{arctan}(\frac{h_i}{d^{hor}_{i,j}})$. $h_i$ is the height of UAV-$i$ from ground level and $d^{hor}_{i,j}$ is the horizontal distant between the UAV-$i$ and the vehicle-$j$.\\
Thus, the probability of non-line-of-sight loss is calculated as:
\begin{equation}
     P^{D2V}_{i,j}(NLoS) = 1- P^{D2V}_{i,j}(LoS)
     \label{eq:pnlos}
\end{equation}
Referring to \cite{art:103-2019}, the path losses with LoS and NLoS links between UAV-$i$ and vehicle-$j$ can be written as:
\begin{align}
    PL^{D2V}_{i,j}(LoS) &= 20\ \text{log}_{10} \left ( \frac{4 \pi f_c d^{euc}_{i,j}}{c} \right ) + \eta_{LoS}\\ 
    PL^{D2V}_{i,j}(NLoS) &= 20\ \text{log}_{10} \left ( \frac{4 \pi f_c d^{euc}_{i,j}}{c} \right ) + \eta_{NLoS} ,
 \label{eq:pathloss}
\end{align}
where $\eta_{LoS}$ and $\eta_{NLoS}$ are the mean additional losses for LoS and NLoS links, $c$ is the speed of light, and $d^{euc}_{i,j} = \sqrt{h_i^2 + \left ( d^{hor}_{i,j} \right ) ^2}$ is the euclidean distance between the UAV-$i$ and vehicle-$j$.\\
As a result, the average path loss of the D2V communication between UAV-$i$ and vehicle-$j$ can be computed as followed:  

\begin{align*}
    PL^{D2V}(i,j) &= P^{D2V}_{i,j}(LoS)PL^{D2V}_{i,j}(LoS) + P^{D2V}_{i,j}(NLoS)PL^{D2V}_{i,j}(NLoS)\\
    &=20 \ \frac{1}{1+a \ \text{exp} \left( -b \left(\theta_{i,j} -a \right) \right)} \left [ \text{log}_{10} \left ( \frac{4 \pi f_c d^{euc}_{i,j}}{c} \right ) + \eta_{LoS} \right ]\\ 
    &+ 20 \ \left(1- P_{i,j}(LoS) \right) \left [ \text{log}_{10} \left ( \frac{4 \pi f_c d^{euc}_{i,j}}{c} \right ) + \eta_{NLoS}\right ]\\
    &= \frac{\eta_{LoS} - \eta_{NLoS}}{1+a \ \text{exp} \left( -b \left(\theta_{i,j} -a \right) \right)}  \\
    &+ 20 \ \text{log}_{10} \left( d^{euc}_{i,j} \text{sec}(\theta_{i,j})\right)  +20 \ \text{log}_{10}\left( \frac{4 \pi f_c}{c} \right) + \eta_{NLoS}
\end{align*}
And the channel gain is given by
\begin{equation}
    G^{D2V}(i,j) = \frac{1}{PL^{D2V}(i,j)}
    \label{eq:channelgain}
\end{equation}
Consequently, the $\text{SNR}_{i,j}$ and the achievable data rate $C_{i,j}$ in bits per second (bps) of the D2V communication are presented in Equation. \eqref{eq:d2vsnr} and \eqref{eq:d2vdatarate} \cite{art:97-2022, art:101-2022}:
\begin{equation}
    \text{SNR}^{D2V}_{i,j} = \frac{p_i^{trans}G^{D2V}_{i,j}}{\sum_{n \in \mathbb{N}_{\text{int}}} p^{trans}_n G^{D2V}_{n,j} + N_0},
    \label{eq:d2vsnr}
\end{equation}
where $p_i^{trans}$ is the transmit power of UAV-$i$, $\mathbb{N}_{\text{int}}$ is the set of possible interfering UAVs, $N_0$ is the noise power.\\
According to Shannon's theorem, the achievable rate of the D2V communication is:
\begin{equation}
    C^{D2V}_{i,j} = B * \text{log}_2 \left(1 + \text{SNR}^{D2V}_{i,j} \right),
    \label{eq:d2vdatarate}
\end{equation}
where $B$ is the bandwidth for the D2V communication. The energy consumption of the UAV transmitting the message is \cite{art:100-2021}:
\begin{equation}
    E^{D2V}_{i,j} = \frac{S_i}{C^{D2V}_{i,j}} \ p_i^{trans},
    \label{eq:transpower}
\end{equation}
where $S_i$ is the size of the message that UAV-$i$ is going to send.

\paragraph{Drone-to-drone path loss}
The D2D path loss is also known as air-to-air path loss. Contrary to the D2V scenario, the D2D path loss is dominated by the free-space LoS propagation. Thus the D2D LoS path loss between UAV-$n$ and UAV-$m$ is given as the same as the D2V communication, as described in Equation. \eqref{eq:agloss}:
\begin{equation}
    PL_{n,m}^{D2D} = 20 \text{log}_{10} \left ( \frac{4 \pi f_c d^{euc}_{n,m}}{c} \right ) + \eta_{LoS}^{D2D},
 \label{eq:agloss}
\end{equation}
where $\eta_{LoS}^{D2D}$ is the mean additional loss of the D2D communication link, $c$ is the speed of light, and $d^{euc}_{n,m} = \sqrt{\left(x_n-x_m\right)^2 + \left(y_n - y_m\right)^2}$ is the euclidean distance between UAV-$n$ and UAV-$m$.\\
Consequently, the channel gain $G_{n,m}^{D2D}$, SNR$_{n,m}^{D2D}$, achievable data rate $C_{n,m}^{D2D}$ and energy consumption $E_{n,m}^{D2D}$ for transmitting a message of size $S_n$ are calculated as followed:
\begin{align}
    G_{n,m}^{D2D} &= \frac{1}{PL_{n,m}^{D2D}}\\
    \text{SNR}_{n,m}^{D2D} &= \frac{p_n G_{n,m}^{D2D}}{\sum_{i \in \mathbb{N}_{\text{int}}} p_i G_{i,j}^{D2D} + N_0}\\
    C_{n,m}^{D2D} &= B * \text{log}_2 \left(1 + \text{SNR}_{n,m}^{D2D} \right)\\
    E_{n,m}^{D2D} &= \frac{S_n}{C_{n,m}^{D2D}} \ p_n^{trans}
    \label{eq:d2denergy}
\end{align}

\subsection{Energy for mobility}
According to \cite{art:DAVNenergy, art:98-2019}, the motion power model $P_i$ of UAV-$i$ of speed $V_i$ is represented as Equation. \eqref{eq:motionenergy}. Thus, the energy consumption to move from location $M(x_1,y_1,z_1)$ to location $M'(x_2,y_2,z_2)$ can be calculated by Equation. \eqref{eq:communicationenergy}:
\begin{equation}
    P_i = P_0 \left ( 1+\frac{3V_i^2}{U_{\text{tip}}^2} \right ) + P_1 \left ( \sqrt{1+\frac{V_i^4}{4v_0^4}} - \frac{V_i^2}{2v_0^2}  \right )^{1/2} + \frac{1}{2} d_0 \rho s A V_i^3 ,
    \label{eq:motionenergy}
\end{equation}
\begin{equation}
    E_i (M, M') = \frac{d_{MM'}}{V_i} P_i 
                = \frac{\sqrt{(x_1-x_2)^2+(y_1-y_2)^2+(z_1-z_2)^2}}{V_i} P_i
    \label{eq:communicationenergy}
\end{equation}
where $P_0$ and $P_i$ are calculated by the following equations as formulated in \cite{art:102-2019}:
\begin{align}
    P_0 &= \frac{\delta}{8} \rho s A \Omega^3 R^3\\
    P_1 &= (1+k) \frac{W^{3/2}}{\sqrt{2 \rho A}}.
    \label{eq:p0p1}
\end{align}
$U_{\text{tip}}$, $\textit{v}_0$,  $d_0$, $s$, $\rho$ and $\mathit{A}$ are constant parameters related to UAV's dynamic properties and air density.


\section{Dataset Generation}
\subsection{Simulation Setup}
\begin{figure}[htbp]
    \centering   
    \includegraphics[width=\textwidth]{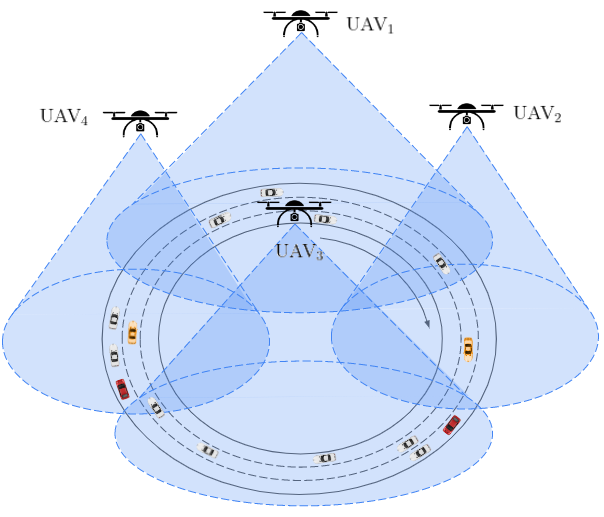}
    \caption{Simulation scenario for dataset generation. Grey cars are ordinary cars, yellow cars are aggressive cars and red cars are emergency cars. The four blues ellipses represent the communication range of each UAV}
    \label{fig:scenario_dataset}
\end{figure}

We consider a 4-km three-lane highway where 4 UAVs hover over the highway, as illustrated in Fig.~\ref{fig:scenario_dataset}. The UAVs move according to the elliptical trajectories as detailed in section \ref{sec:elliptical}. The maximum horizontal speed of UAVs is 30 km/h, the maximum vertical speed is 10 km/h, the limited flying height is between 100 m - 150 m \cite{art:45-2017,art:uavspeed-2022}. \\ 
On the other hand, vehicle trajectories are retrieved by the TraCI interface of Simulation of Urban MObility (SUMO) \cite{art:traci, art:sumo}, a free and open source traffic simulator. Vehicles move according to the Krauss mobility model and LC2013 lane change model. The maximum allowed velocity is 100 km/h. For a more authentic scenario, some vehicles are set to be “aggressive” with impolite behaviors such as low intention to cooperate with others, stay in the leftmost lane for a long time and exceed the speed limit, represented by the yellow cars. Moreover, some vehicles are ambulances that have higher priority to pass and lead the in front vehicles to initiate lane change, represented by the red cars \cite{art:hjl-2022}.

\subsection{UAV Trajectory}\label{sec:elliptical}
\subsubsection{Horizontal trajectory for x and y}
\begin{figure}[htbp]
\centering
\begin{minipage}[t]{0.4\textwidth}
\centering
\includegraphics[width=\textwidth]{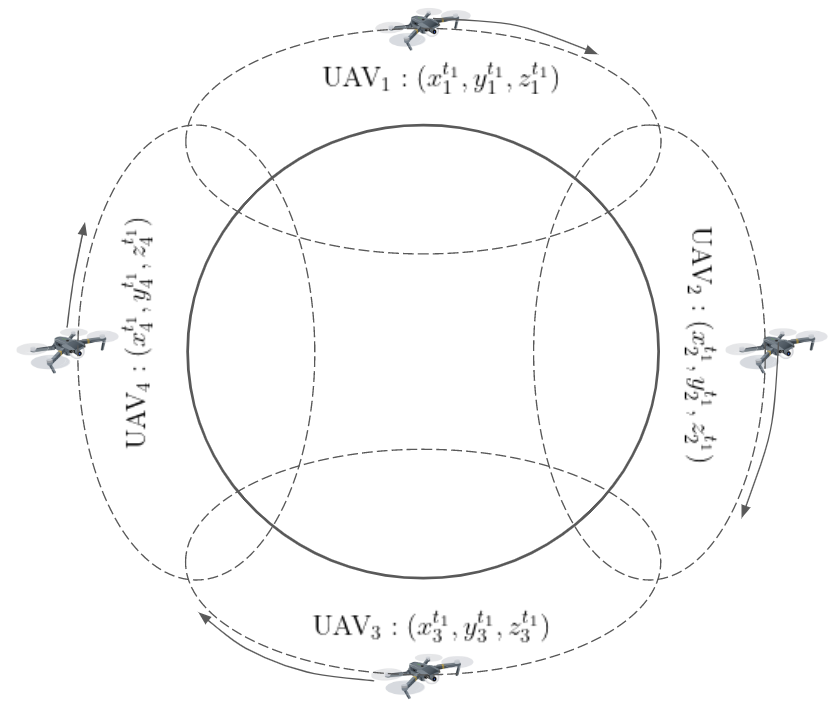}
\end{minipage}
\begin{minipage}[t]{0.4\textwidth}
\centering
\includegraphics[width=\textwidth]{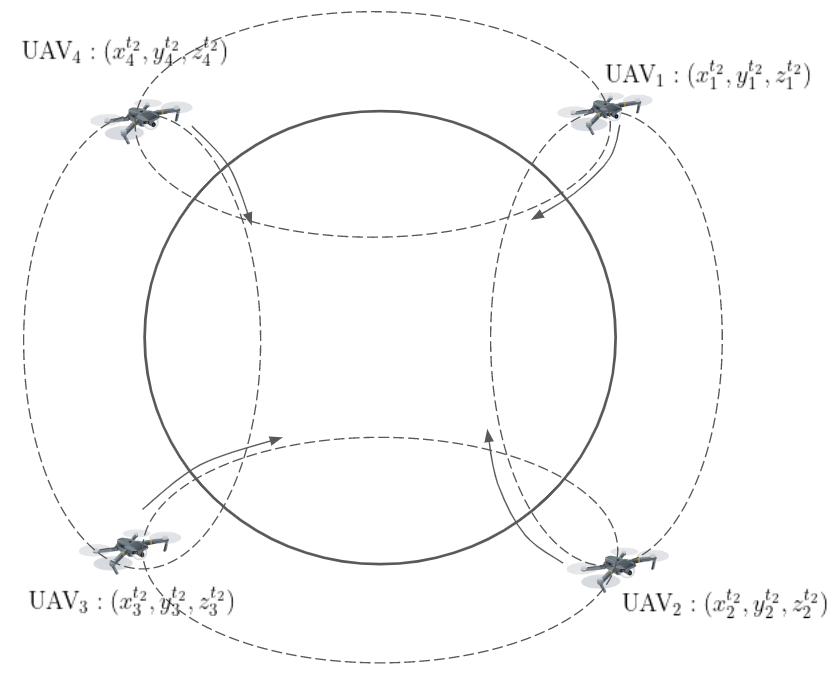}
\end{minipage}

\begin{minipage}[t]{0.4\textwidth}
\centering
\includegraphics[width=\textwidth]{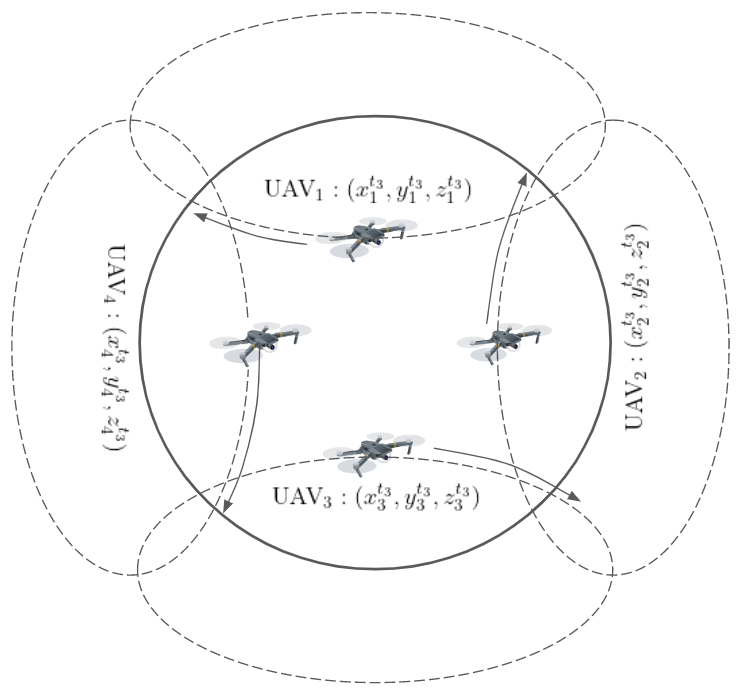}
\end{minipage}
\begin{minipage}[t]{0.4\textwidth}
\centering
\includegraphics[width=\textwidth]{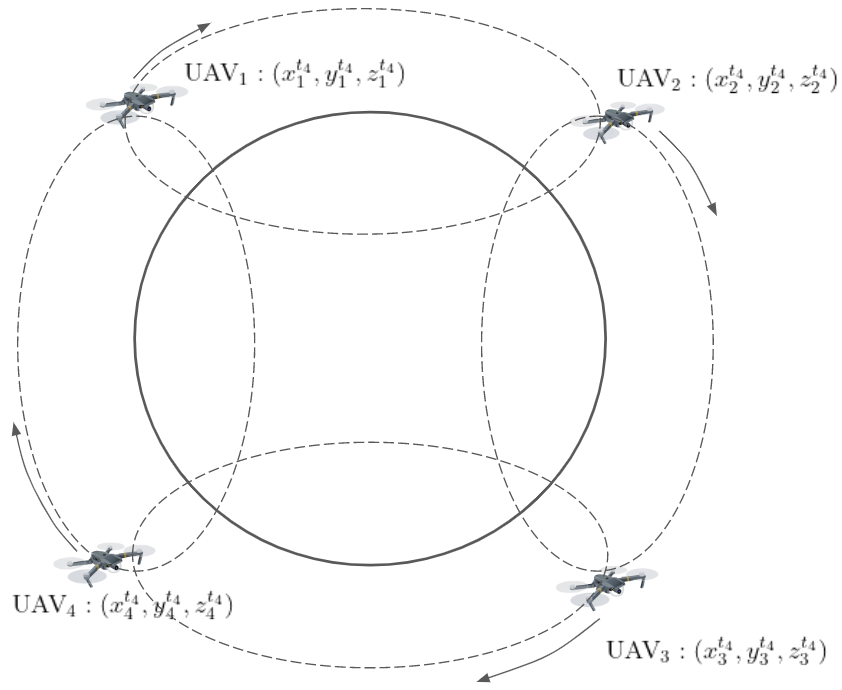}
\end{minipage}
\caption{Examples of the generated trajectories of UAVs at time $t_1$, $t_2$, $t_3$ and $t_4$}
\label{fig:elliptic}
\end{figure}

We adopt elliptical trajectory for the four UAVs as illustrated in Fig. \ref{fig:elliptic}. The UAVs move according to the elliptic curve. The initial mathematical expression is shown in Equation. \eqref{eq:elliptic}:
\begin{equation}
    \frac{x^2}{a^2} + \frac{y^2}{b^2} = 1,  
    \label{eq:elliptic}
\end{equation}
where 2a is the width of the ellipse and 2b is the height of the ellipse.

\begin{figure}[htpb]
    \centering
    \includegraphics[scale=0.9]{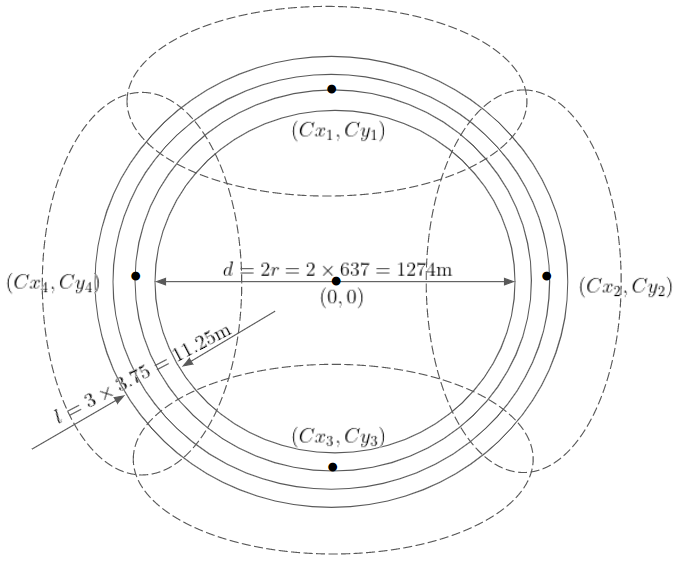}
    \caption{Details of the ellipses}
    \label{fig:center}
\end{figure}

In our simulation scenario, the four ellipses are distributed on a circle centered on $(0,0)$ with a radius of 637 meters, representing the circular highway, as shown in Fig. \ref{fig:center}. Consequently, $2a=637 \times 2=1274$, $2b=637$. The horizontal speed and vertical speed of the drones are liimited to 30km/h ($\approx$ 8.33m/s) and 10km/h ($\approx$ 2.78m/s). The traffic state is updated every 0.4 second. Thus, x and y positions are functions of simulation step, $s$, where $v_{hor}$ and $v_{vrt}$ are the horizontal speed and vertical speed in meter per second: 
\begin{align*}
 x &= v_{hor} \times 0.4s\\ 
 y &= v_{vrt} \times 0.4s
\end{align*}
It should be noted that the speed of the four drones are set to 5, 10, 20, 30 km/h. 
On the other hand, the center coordinates of the four UAVs are 
\begin{align*}
 \left( Cx_1, Cy_1 \right) &= (0, 637)\\ 
 \left( Cx_2, Cy_2 \right) &= (637, 0)\\
 \left( Cx_3, Cy_3 \right) &= (0, -637)\\
  \left( Cx_4, Cy_4 \right) &= (-637, 0)\\
\end{align*}
Finally, the trajectories with expression in the form of 
$$\frac{(x-Cx_i)^2}{a^2} + \frac{(y-Cy_i)^2}{b^2} =1$$ 
for the four UAVs can be represented as in the following equations:
\begin{align*}
 \frac{x^2}{a^2} + \frac{(y-637)^2}{b^2} & =1\\
 \frac{(x-637)^2}{a^2} + \frac{y^2}{b^2} & =1\\
 \frac{x^2}{a^2} + \frac{(y+637)^2}{b^2} & =1\\
 \frac{(x+637)^2}{a^2} + \frac{y^2}{b^2} & =1
\end{align*}

\subsubsection{Vertical trajectory for z}
We adopt a random walk trajectory to determine $z_i, i\in [1,4]$. It should be noticed that at each step, the UAV moves up or down according to the predefined probability array $p=(p_1,1-p_1)$, where $p_1$ is the probability to moving up, and $p_2$ is the probability to moving down. In the simulation, we set $p=(0.5,0.5)$. The random walk algorithm is detailed in Algorithm. \ref{algo:randomwalk}.
\begin{algorithm}[htbp]
\caption{\texttt{Random Walk Algorithm for Vertical Trajectory Generation}}
\label{algo:randomwalk}
\begin{algorithmic}[1]
    \State \textbf{Input:} $S$ is the number of simulation steps, $Z$ is the vertical action range, $p$ and $1-p$ are the probability to go up and down at each step, $\Delta z$ is the vertical step length
    \State Initialize vertical trajectory as an all-zero array $\boldsymbol{z}$ of length $S$
    \For{$i \leftarrow$ 0 to $S$}
        \State Randomly choose a vertical direction $up$ or $down$ according to $p$
        \If{the chosen direction is $up$ (or $down$)}
            \If {$\boldsymbol{z}[i] + \Delta z (or -\Delta z) \in Z$}
                \State Perform the movement
            \EndIf 
        \EndIf
    \EndFor
\end{algorithmic}
\end{algorithm}

\subsubsection{Final trajectories}
\begin{figure}[htbp]
\centering
\begin{minipage}[t]{0.4\textwidth}
\centering
\includegraphics[width=\textwidth]{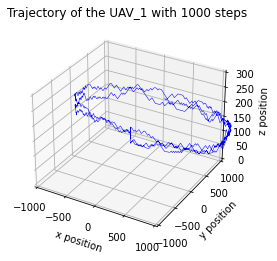}
\end{minipage}
\begin{minipage}[t]{0.4\textwidth}
\centering
\includegraphics[width=\textwidth]{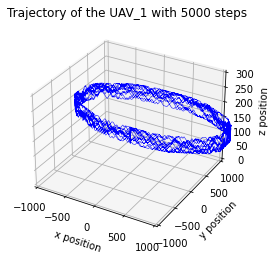}
\end{minipage}

\begin{minipage}[t]{0.4\textwidth}
\centering
\includegraphics[width=\textwidth]{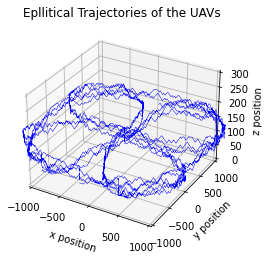}
\end{minipage}
\begin{minipage}[t]{0.4\textwidth}
\centering
\includegraphics[width=\textwidth]{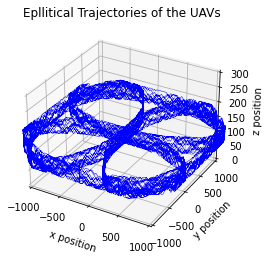}
\end{minipage}
\caption{Elliptical trajectories of UAVs}
\label{fig:ellp_traj}
\end{figure}

Examples of the generated trajectories of the UAVs are shown in Fig. \ref{fig:ellp_traj}.

\subsection{Dataset parameters}
The dataset will be generated through a branch of simulations with different vehicular density $\rho$. The samples (i.e. the vehicles and drones kinematic parameters) are retrieved, computed and stored every 0.4s. The parameters used for simulation are shown in Table. \ref{tab:notations}. 
Each sample consists of the observations of each of the 4 UAVs, as well as the overall collision rate on the highway, as represented by the following:
$$
\mathnormal{o}= \left \{ \mathnormal{o_{1}}, \mathnormal{o_{2}}, \mathnormal{o_{3}}, \mathnormal{o_{4}}, \bar{r} \right \},$$
where $\mathnormal{o_{1}}, \mathnormal{o_{2}}, \mathnormal{o_{3}}$, and $\mathnormal{o_{4}}$ are the observations of UAV$_1$,UAV$_2$,UAV$_3$ and UAV$_4$, respectively. $\bar{r}$ is the total collision number on the highway.\\
An observation of UAV-$i$ at step $t$ is expressed as
\begin{equation}
    \mathnormal{o_{i}}[t]= \left \{x_i[t], y_i[t], z_i[t], \theta_i[t], \rho_i, \bar{W_i}, \bar{E_i}, \bar{t_{ri}}, \bar{t_{bi}} \right \},
    \label{eq:observation}
\end{equation}
where $x_i[t]$ is the longitudinal position of UAV-$i$ at step $t$, $y_i[t]$ denotes the lateral position at step $t$, $z_i[t]$ is the height at step $t$, $\theta_i[t]$ is the heading direction at step $t$ ($\theta_i[t]=1$ means moving up, $\theta_i[t]$=-1 means moving down), $\rho_i$ is the number of vehiculars in the communication range of UAV-$i$, $\bar{W_i}$ the mean waiting time of the vehicle requests in the communication range of UAV-$i$, $\bar{E_i}$ the average energy consumption of UAV-$i$, $\bar{t_{ri}}$ and $\bar{t_{bi}}$ the mean risky time and mean blocking time of the vehicles in the communication range of the UAV-$i$. The algorithm for generating the dataset is presented in Algorithm. \ref{algo:dataset}\\
\begin{algorithm}[htbp]
\caption{\texttt{Algorithm for dataset generation}}
\label{algo:dataset}
\begin{algorithmic}[1]
    \State \textbf{Input:} $S$ is the number of simulation steps, $\mathbb{V}$ is the set of vehicles on the road
    \State Initialize
    \For{$t \leftarrow$ 0 to $S$}
        \For{every vehicle $j$, $j \in \mathbb{V}$}
            \State Determine the corresponding UAV of the vehicle according to the vehicle's GPS position and UAV's coverage, denote the corresponding UAV as UAV-$i$
            \State Compute $W_i^j$, $E_i^j$ according to Equation. \eqref{eq:delay} and Equation.  \eqref{eq:energyconsumption}
            \State Retrieve $t_{ri}^j$, $t_{bi}^j$ from vehicle state information
            \State Store $W_i^j$, $E_i^j$, $t_{ri}^j$ and $t_{bi}^j$ in the UAV-$i$'s buffer
        \EndFor
        \For{every UAV $i$, $i \in [1,4]$}
            \State Compute the average $\bar{W_i}$, $\bar{E_i}$, $\bar{t_{ri}}$, $\bar{t_{bi}}$
            \State Store current position $x_i[t]$, $y_i[t]$, $z_i[t]$, current heading direction $\theta_i[t]$, current vehicular density $\rho_i$, and $\bar{W_i}$, $\bar{E_i}$, $\bar{t_{ri}}$, $\bar{t_{bi}}$ as current observation of UAV-$i$
        \EndFor
        \State Store $\bar{r}$, number of collisions happen at the current step
    \EndFor
\end{algorithmic}
\end{algorithm}

\section{Notations and Terminologies}
The notations and terminologies used in this report are summarized in Table \ref{tab:notations}.\\
The drones are equipped with SoC semiconductors: Snapdragon 821 with Quad-core up to 2.15GHz. The mean service time for the two types of messages are calculated as followed: 
\begin{itemize}
    \item The length of safety message is exponentially distributed with parameter $\lambda_1' = \frac{2.15\text{GHz}\times 4}{512\times 8\text{bits}} \approx 2.1 \times 10^6$. Thus, $E[B_1] = \frac{1}{\lambda_1'}=4.763 \times 10^{-7}$ s, $E[B_1^2] = \frac{1}{(\lambda_1')^{2}} \approx 2.269 \times 10^{-13}$. 
    \item On the other hand, the length of vehicle state information is a constant and equals to 32 bytes. Thus, $E[B_2] = \frac{32 \times 8}{2.15\text{GHz} \times 4}=2.977 \times 10^{-8}$ s, $E[B_2^2] = 0$.
\end{itemize}

{
\footnotesize
\begin{longtable}[c]{|c|c|c|c|c|}
\caption{MAIN NOTATIONS AND TERMINOLOGIES}
\label{tab:notations}\\
\hline
\textbf{Section} & \textbf{Parameter} & \textbf{Meaning} & \textbf{Value} & \textbf{Reference} \\ \hline
\endfirsthead
\multicolumn{5}{c}%
{{\bfseries Table \thetable\ continued from previous page}} \\
\hline
\textbf{Section} & \textbf{Parameter} & \textbf{Meaning} & \textbf{Value} & \textbf{Reference} \\ \hline
\endhead
 & $E[W_{i,j}^{D2V}]$ & The D2V propagation delays & $\frac{d^{'}_{i,j}}{r_i}$ &  \\ \cline{2-5} 
 & $E[W_{i,j}^{V2D}]$ & The V2D propagation delays & $\frac{d_{i,j}}{r_i}$ &  \\ \cline{2-5} 
\begin{tabular}[c]{@{}c@{}}V2D and D2V\\ \\ Propagation\\ \\ Delay\end{tabular} & \begin{tabular}[c]{@{}c@{}}$d_{i,j}$ and \\ $d^{'}_{i,j}$\end{tabular} & \begin{tabular}[c]{@{}c@{}}The distances between \\ the UAV and the vehicle-$j$ \\ from which the request is\\  sent during the V2D \\ and D2V communication\end{tabular} & $\sqrt{(x_i-x_j)^2-(y_i-y_j)^2}$ &  \\ \cline{2-5} 
 & $r_{i,j}$ & \begin{tabular}[c]{@{}c@{}}The achievable \\ propagation rate  of \\ the communication lnk\end{tabular} & $B * \text{log}_2 \left(1 + \text{SNR}_{i,j} \right)$ &  \\ \hline
 & $\lambda _i$ & \begin{tabular}[c]{@{}c@{}}The arrival rate of a \\ class-$i$ request\end{tabular} & \begin{tabular}[c]{@{}c@{}}$\lambda_1=20\%$ of vehicle number\\ $\lambda_2=2.5$\end{tabular} & \cite{art:22-2017} \\ \cline{2-5} 
 & $\mu$ & UAV service rate & 250 & \cite{art:90-2014} \\ \cline{2-5} 
 & $E[B _i]$ & \begin{tabular}[c]{@{}c@{}}The mean service time\\  at the drone of a \\ class-$i$ request\end{tabular} & \begin{tabular}[c]{@{}c@{}}$E[B_1]=4.763 \times 10^{-7}$, \\$E[B_1^2]=2.269 \times 10^{-13}$\\ $E[B_2]=2.977 \times 10^{-8}$, \\$E[B_2^2]=0$\end{tabular}  & \cite{url:Snapdragon821} \\ \cline{2-5} 
Queuing & $E[W _i]$ & \begin{tabular}[c]{@{}c@{}}The mean waiting time \\ in the queue of a \\ class-$i$ request\end{tabular} &  &  \\ \cline{2-5} 
Delay & $E[R _i]$ & \begin{tabular}[c]{@{}c@{}}The mean residual service\\  time of a class-$i$ request\end{tabular} &  &  \\ \cline{2-5} 
at the & $E[S _i]$ & \begin{tabular}[c]{@{}c@{}}The mean sojourn time\\  of a class-$i$ request\end{tabular} & $E[S _i] = E[B _i] + E[W _i]$ &  \\ \cline{2-5} 
Drone's & $E[L _i]$ & \begin{tabular}[c]{@{}c@{}}The average number of \\ requests of class $i$ \\ waiting in the queue\end{tabular} &  &  \\ \cline{2-5} 
Side & $E[W_{i,j}]$ & \begin{tabular}[c]{@{}c@{}}The total V2D delay of a \\ class-$i$ request from \\ a vehicle-$j$\end{tabular} &  &  \\ \cline{2-5} 
 & $W_{i,j}^{V2D}$ & V2D propagation delay &  &  \\ \cline{2-5} 
 & $W_{i,j}^{D2V}$ & \begin{tabular}[c]{@{}c@{}}Drone-to-vehicle (D2V)\\  propagation delay\end{tabular} &  &  \\ \cline{2-5} 
 & $W_{i,j}^s$ & \begin{tabular}[c]{@{}c@{}}Queuing delay at \\ the drone side\end{tabular} &  &  \\ \cline{2-5} 
 & $T_i$ & \begin{tabular}[c]{@{}c@{}}The maximum waiting \\ time for request class-$i$\end{tabular} & 0.2 s & \cite{art:91-2021} \\ \hline
 & $a$ and $b$ & \begin{tabular}[c]{@{}c@{}}Environmental constant \\ depending on rural\\  or urban area\end{tabular} & \begin{tabular}[c]{@{}c@{}}14.39\\ 0.13\end{tabular} & \cite{art:97-2022} \\ \cline{2-5} 
 & $h_i$ & \begin{tabular}[c]{@{}c@{}}The height of UAV-$i$ \\ from ground level\end{tabular} & [100,150] & \cite{art:45-2017} \\ \cline{2-5} 
 & $\theta_{i,j}$ & \begin{tabular}[c]{@{}c@{}}Elevation angle between \\ UAV-$i$ and vehicle-$j$\end{tabular} & $\text{arctan}(\frac{h_i}{d^{hor}_{i,j}})$ &  \\ \cline{2-5} 
 & $d^{hor}_{i,j}$ & \begin{tabular}[c]{@{}c@{}}The horizontal distance \\ between the UAV$_i$\\  and the vehicle$_j$\end{tabular} & $\sqrt{(x_1-x_2)^2+(y_1-y_2)^2}$ &  \\ \cline{2-5} 
 & $d^{euc}_{i,j}$ & \begin{tabular}[c]{@{}c@{}}The euclidean distance \\ between the UAV-$i$ \\ and vehicle-$j$\end{tabular} & $\sqrt{h_i^2 + \left ( d^{hor}_{i,j} \right ) ^2}$ &  \\ \cline{2-5} 
 & $P_{i,j}(LoS)$ & \begin{tabular}[c]{@{}c@{}}The probability of\\  line-of-sight link\end{tabular} & $P_{i,j}(LoS) = \frac{1}{1+a \ \text{exp} \left( -b \left(\theta_{i,j} -a \right) \right)}$ &  \\ \cline{2-5} 
 & $P_{i,j}(NLoS)$ & \begin{tabular}[c]{@{}c@{}}The probability of \\ non-line-of-sight link\end{tabular} & $1-P_{i,j}(LoS)$ &  \\ \cline{2-5} 
Energy & $PL_{i,j}(LoS)$ & Path loss with LoS  link & $20\ \text{log} \left ( \frac{4 \pi f_c d^{euc}_{i,j}}{c} \right ) + \eta_{LoS}$ &  \\ \cline{2-5} 
for D2V & $PL_{i,j}(NLoS)$ & Path loss with NLoS link & $20\ \text{log} \left ( \frac{4 \pi f_c d^{euc}_{i,j}}{c} \right ) + \eta_{NLoS}$ &  \\ \cline{2-5} 
\begin{tabular}[c]{@{}c@{}}Communi-\\ \\ cations\end{tabular} & $f_c$ & \begin{tabular}[c]{@{}c@{}}Transmit frequency for \\ uplink and downlink of\\ the D2V communication\end{tabular} & 2.4GHz & \cite{art:45-2017} \\ \cline{2-5} 
 & \begin{tabular}[c]{@{}c@{}}$\eta_{LoS}$ and\\  $\eta_{NLoS}$\end{tabular} & \begin{tabular}[c]{@{}c@{}}Additional losses for \\ LoS and NLoS links\end{tabular} & \begin{tabular}[c]{@{}c@{}}1 dB\\ 20 dB\end{tabular} & \cite{art:97-2022} \\ \cline{2-5} 
 & $c$ & The speed of light & $3 \times 10^8$ m/s &  \\ \cline{2-5} 
 & $G(i,j)$ & \begin{tabular}[c]{@{}c@{}}The channel gain of the \\ communication link between\\  the UAV-$i$ and vehicle-$j$\end{tabular} & $G(i,j) = \frac{1}{PL(i,j)}$ &  \\ \cline{2-5} 
 & $\text{SNR}_{i,j}$ & The signal-to-noise ratio & $\frac{p_iG_{i,j}}{\sum_{n \in \mathbb{N}_{\text{int}}} p_n G_{n,j} + N_0}$ &  \\ \cline{2-5} 
 & $C_{i,j}$ & \begin{tabular}[c]{@{}c@{}}The achievable data rate \\ in bits per second (bps)\end{tabular} & $B \cdot \text{log}_2 \left(1 + \text{SNR}_{i,j} \right)$ &  \\ \cline{2-5} 
 & $p_i^{trans}$ & \begin{tabular}[c]{@{}c@{}}The transmit power \\ of UAV-$i$\end{tabular} & 280 mW & \cite{art:45-2017} \\ \cline{2-5} 
 & $N_0$ & The noise power & -174 dB/Hz & \cite{art:104-2019} \\ \cline{2-5} 
 & $B$ & \begin{tabular}[c]{@{}c@{}}The bandwidth for the \\ D2V communication\end{tabular} & 100 MHz & \cite{art:45-2017} \\ \cline{2-5} 
 & $S_i$ & \begin{tabular}[c]{@{}c@{}}The size of the message that \\ UAV-$i$ is going to send\end{tabular} & 512 bytes & \cite{art:45-2017} \\ \hline
 & $PL_{n,m}^{D2D}$ & \begin{tabular}[c]{@{}c@{}}The D2D LoS path loss\\  between UAV-$n$\\ and UAV-$m$\end{tabular} & $20 \text{log} \left ( \frac{4 \pi f_c d^{euc}_{n,m}}{c} \right ) + \eta_{LoS}^{D2D}$ &  \\ \cline{2-5} 
 & $\eta_{LoS}^{D2D}$ & \begin{tabular}[c]{@{}c@{}}The mean additional \\ loss of the D2D \\ communication link\end{tabular} &  &  \\ \cline{2-5} 
\begin{tabular}[c]{@{}c@{}}Energy\\ \\ for D2D\end{tabular} & $d^{euc}_{n,m}$ & \begin{tabular}[c]{@{}c@{}}The euclidean distance \\ between UAV-$n$ \\ and UAV-$m$\end{tabular} & $\sqrt{\left(x_n-x_m\right)^2 + \left(y_n - y_m\right)^2}$ &  \\ \cline{2-5} 
Communi- & $G_{n,m}^{D2D}$ & The channel gain & $\frac{1}{PL_{n,m}^{D2D}}$ &  \\ \cline{2-5} 
cations & $SNR_{n,m}^{D2D}$ & The signal-to-noise ratio & $\frac{p_n G_{n,m}^{D2D}}{\sum_{i \in \mathbb{N}_{\text{int}}} p_i G_{i,j}^{D2D} + N_0}$ &  \\ \cline{2-5} 
 & $C_{n,m}^{D2D}$ & Achievable data rate & $B \cdot \text{log}_2 \left(1 + \text{SNR}_{n,m}^{D2D} \right)$ &  \\ \cline{2-5} 
 & $E_{n,m}^{D2D}$ & \begin{tabular}[c]{@{}c@{}}Energy consumption for \\ transmitting a message\\  of size $S_n$\end{tabular} & $\frac{S_n}{C_{n,m}^{D2D}} \ p_n^{trans}$ &  \\ \cline{2-5} 
 & $S_n$ & The transmitted message size & 512 bytes & \cite{art:45-2017} \\ \hline
 & $V_i$ & \begin{tabular}[c]{@{}c@{}}Horizontal speed of UAV-$i$\\ Vertical speed of UAV-$i$\end{tabular} & \begin{tabular}[c]{@{}c@{}}30 km/h\\ 10 km/h\end{tabular} &  \\ \cline{2-5} 
 & $P_0$ & \begin{tabular}[c]{@{}c@{}}Power constant representing\\  the blade profile\end{tabular} & 84.14 $N$ & \cite{art:102-2019} \\ \cline{2-5} 
Energy & $P_1$ & \begin{tabular}[c]{@{}c@{}}Power constant representing\\  the induced power levels\\  in hovering status\end{tabular} & 88.63 $N$ & \cite{art:102-2019} \\ \cline{2-5} 
for & $U_{\text{tip}}$ & The tip speed of the rotor blade & 120 m/s & \cite{art:102-2019} \\ \cline{2-5} 
Mobility & $\textit{v}_0$ & \begin{tabular}[c]{@{}c@{}}The mean rotor induced \\ velocity in hover\end{tabular} & 4.03 & \cite{art:102-2019} \\ \cline{2-5} 
 & $d_0$ & The fuselage drag ratio & 0.6 & \cite{art:102-2019} \\ \cline{2-5} 
 & $s$ & Rotor solidity & 0.05 & \cite{art:102-2019} \\ \cline{2-5} 
 & $\rho$ & Air density & 1.225 kg/m$^3$ & \cite{art:102-2019} \\ \cline{2-5} 
 & $\mathit{A}$ & Rotor disc area & 0.503 m$^2$ & \cite{art:102-2019} \\ \hline
\end{longtable}

}

\section{Acknowledgement}
This work is part of the Ph.D work funded by DigiCosme project: ANR11LABEX0045DIGICOSME, supervised by Djamal Zeghlache (Télécom SudParis) and Rola Naja (ECE Paris).


\begin{thebibliography}{00}
\bibitem{art:45} W. Shi, H. Zhou, J. Li, W. Xu, N. Zhang and X. Shen, "Drone Assisted Vehicular Networks: Architecture, Challenges and Opportunities," in IEEE Network, vol. 32, no. 3, pp. 130-137, May/June 2018, doi: 10.1109/MNET.2017.1700206.

\bibitem{art:hjl-2022} J. Hao, R. Naja, and D. Zeghlache, ``Drone-assisted lane change maneuver using reinforcement learning with dynamic reward function," IEEE WiMob, pp. 319-325, 2022.

\bibitem{art:atgloss} A. Al-Hourani, S. Kandeepan and S. Lardner, "Optimal LAP Altitude for Maximum Coverage," in IEEE Wireless Communications Letters, vol. 3, no. 6, pp. 569-572, Dec. 2014, doi: 10.1109/LWC.2014.2342736.

\bibitem{art:94-2021} Raza A, Bukhari SHR, Aadil F, Iqbal Z. An UAV-assisted VANET architecture for intelligent transportation system in smart cities. International Journal of Distributed Sensor Networks. 2021;17(7). doi:10.1177/15501477211031750

\bibitem{art:DAVNenergy} Y. Zeng, J. Xu and R. Zhang, "Energy Minimization for Wireless Communication With Rotary-Wing UAV," in IEEE Transactions on Wireless Communications, vol. 18, no. 4, pp. 2329-2345, April 2019, doi: 10.1109/TWC.2019.2902559.

\bibitem{art:98-2019} H. Ghazzai, A. Khattab and Y. Massoud, "Mobility and Energy Aware Data Routing for UAV-Assisted VANETs," 2019 IEEE International Conference on Vehicular Electronics and Safety (ICVES), 2019, pp. 1-6, doi: 10.1109/ICVES.2019.8906323.

\bibitem{art:103-2019} C. -C. Lai, C. -T. Chen and L. -C. Wang, "On-Demand Density-Aware UAV Base Station 3D Placement for Arbitrarily Distributed Users With Guaranteed Data Rates," in IEEE Wireless Communications Letters, vol. 8, no. 3, pp. 913-916, June 2019, doi: 10.1109/LWC.2019.2899599.

\bibitem{art:97-2022} S. Mokhtari, N. Nouri, J. Abouei, A. Avokh and K. N. Plataniotis, "Relaying Data With Joint Optimization of Energy and Delay in Cluster-Based UAV-Assisted VANETs," in IEEE Internet of Things Journal, vol. 9, no. 23, pp. 24541-24559, 1 Dec.1, 2022, doi: 10.1109/JIOT.2022.3188563.

\bibitem{art:101-2022} Tran, Dinh-Hieu, Symeon Chatzinotas, and Björn Ottersten. "Throughput Maximization for Backscatter-and Cache-Assisted Wireless Powered UAV Technology." IEEE Transactions on Vehicular Technology 71.5 (2022): 5187-5202.

\bibitem{art:100-2021} Wu, Gaoxiang, et al. "Adaptive Edge Caching in UAV-assisted 5G Network." 2021 IEEE Global Communications Conference (GLOBECOM). IEEE, 2021.

\bibitem{art:rate} R. Yu, J. Ding, X. Huang, M. -T. Zhou, S. Gjessing and Y. Zhang, "Optimal Resource Sharing in 5G-Enabled Vehicular Networks: A Matrix Game Approach," in IEEE Transactions on Vehicular Technology, vol. 65, no. 10, pp. 7844-7856, Oct. 2016, doi: 10.1109/TVT.2016.2536441.

\bibitem{art:45-2017} W. Shi, H. Zhou, J. Li, W. Xu, N. Zhang and X. Shen, "Drone Assisted Vehicular Networks: Architecture, Challenges and Opportunities," in IEEE Network, vol. 32, no. 3, pp. 130-137, May/June 2018, doi: 10.1109/MNET.2017.1700206.

\bibitem{art:uavspeed-2022} AMA Style
Wubben J, Morales C, Calafate CT, Hernández-Orallo E, Cano J-C, Manzoni P. Improving UAV Mission Quality and Safety through Topographic Awareness. Drones. 2022; 6(3):74. https://doi.org/10.3390/drones6030074

\bibitem{art:90-2014} S. Fowler, C. H. Häll, D. Yuan, G. Baravdish and A. Mellouk, "Analysis of vehicular wireless channel communication via queueing theory model," 2014 IEEE International Conference on Communications (ICC), 2014, pp. 1736-1741, doi: 10.1109/ICC.2014.6883573.

\bibitem{url:Snapdragon821} https://www.modalai.com/pages/snapdragon-flight

\bibitem{art:91-2021} Jia, Zehan, et al. "Learning-based queuing delay-aware task offloading in collaborative vehicular networks." ICC 2021-IEEE International Conference on Communications. IEEE, 2021.

\bibitem{art:102-2019} Y. Zeng, J. Xu and R. Zhang, "Energy Minimization for Wireless Communication With Rotary-Wing UAV," in IEEE Transactions on Wireless Communications, vol. 18, no. 4, pp. 2329-2345, April 2019, doi: 10.1109/TWC.2019.2902559.

\bibitem{art:22-2017} Oubbati, Omar Sami, et al. "Intelligent UAV-assisted routing protocol for urban VANETs." Computer communications 107 (2017): 93-111.

\bibitem{art:104-2019} Zhou, Conghao, et al. "Delay-aware IoT task scheduling in space-air-ground integrated network." 2019 IEEE Global Communications Conference (GLOBECOM). IEEE, 2019.

\bibitem{art:traci} Wegener, A., Piórkowski, M., Raya, M., Hellbrück, H., Fischer, S., Hubaux, J. P.: TraCI: an interface for coupling road traffic and network simulators. In: 11th communications and networking simulation symposium, pp. 155--163, ACM, USA (2008)

\bibitem{art:sumo} Lopez, P. A., Behrisch, M., Bieker-Walz, L., Erdmann, J., Flötteröd, Y. P., Hilbrich, R., ..., Wießner, E.: Microscopic traffic simulation using sumo. In: 21st International Conference on Intelligent Transportation Systems (ITSC), pp. 2575--2582, IEEE, USA (2018)

\bibitem{art:utsim} 1.
Al-Mousa A, Sababha BH, Al-Madi N, Barghouthi A, Younisse R. UTSim: A framework and simulator for UAV air traffic integration, control, and communication. International Journal of Advanced Robotic Systems. 2019;16(5). doi:10.1177/1729881419870937

\bibitem{art:avens} SCHOABA, V. ; SIKANSI, F. E. G. ; PIGATTO, D. F. ; BRANCO, K. R. L. J. C. ; BRANCO, L. C. . Digital Signature for Mobile Devices: A New Implementation and Evaluation. International Journal of Future Generation Communication and Networking, v. 4, p. 23-36, 2011.

\bibitem{url:randomwalk} geeksforgeeks-random-walk-implementation-python \url{https://www.geeksforgeeks.org/random-walk-implementation-python/}, 2017-10-22.


\end{thebibliography}
\end{document}